\documentclass[runningheads]{llncs}
\usepackage{graphicx}

\usepackage{tikz}
\usepackage{comment}
\usepackage{amsmath,amssymb} 
\usepackage{color}
\usepackage{booktabs}
\usepackage[pagebackref,breaklinks,colorlinks]{hyperref}

\usepackage[accsupp]{axessibility}  

\begin{document}
\pagestyle{headings}
\mainmatter
\def\ECCVSubNumber{561}  

\title{Particle Video Revisited: Tracking Through Occlusions Using Point Trajectories}

\titlerunning{Particle Video Revisited}
%
\author{Adam W. Harley \and
Zhaoyuan Fang \and
Katerina Fragkiadaki}
\authorrunning{A. W. Harley et al.}
%
\institute{Carnegie Mellon University\\
\email{\{aharley, zhaoyuaf, katef\}@cs.cmu.edu}\\
Project page: \url{https://particle-video-revisited.github.io}
}
\maketitle

\newcommand\adam[1]{\textcolor{magenta}{#1}}
\newcommand\todo[1]{\textcolor{red}{#1}}
\newcommand\red[1]{\textcolor{red}{#1}}
\newcommand\gist[1]{\textcolor{cyan}{#1}}
\newcommand\teal[1]{\textcolor{teal}{#1}}
\newcommand\pink[1]{\textcolor{magenta}{#1}}

\newcommand{\M}{\mathcal{M}} 
\newcommand{\R}{\mathcal{R}} 
\newcommand{\Mt}{\mathcal{M}^{(t)}} 
\newcommand{\Mz}{\mathcal{M}^{(0)}} 
\newcommand{\Mo}{\mathcal{M}^{(1)}} 
\newcommand{\Mi}{\mathcal{M}^{(i)}} 
\newcommand{\Mj}{\mathcal{M}^{(j)}} 
\newcommand{\Mto}{\mathcal{M}^{(t+1)}} 

\newcommand{\Tt}{\mathcal{T}^{(t)}} 
\newcommand{\Tz}{\mathcal{T}^{(0)}} 
\newcommand{\To}{\mathcal{T}^{(1)}} 

\newcommand{\I}{I} 
\newcommand{\D}{D} 

\newcommand{\loss}{\mathcal{L}}

\newcommand{\Ot}{{O}^{(t)}}
\newcommand{\Oz}{{O}^{(1)}}
\newcommand{\Ct}{{O}^{(t)}}
\newcommand{\Cht}{\hat{{O}}^{(t)}}
\newcommand{\Iht}{\hat{{I}}^{(t)}}

\begin{abstract}

Tracking pixels in videos is typically studied as an optical flow estimation problem, where every pixel is described with a displacement vector that locates it in the next frame. Even though wider temporal context is freely available, prior efforts to take this into account have yielded only small gains over 2-frame methods. In this paper, we revisit Sand and Teller's ``particle video'' approach, and study pixel tracking as a long-range motion estimation problem, where every pixel is described with a trajectory that locates it in multiple future frames. We re-build this classic approach using components that drive the current state-of-the-art in flow and object tracking, such as dense cost maps, iterative optimization, and learned appearance updates. We train our models using long-range amodal point trajectories mined from existing optical flow data that we synthetically augment with multi-frame occlusions. We test our approach in trajectory estimation benchmarks and in keypoint label propagation tasks, and compare favorably against state-of-the-art optical flow and feature tracking methods.

\end{abstract}

\section{Introduction} \label{sec:intro}

In 2006, Sand and Teller~\cite{particlevideo} wrote that there are two dominant approaches to motion estimation in video: feature matching and optical flow. This is still true today. 
In their paper, they proposed a new motion representation called a ``particle video'', which they presented as a middle-ground between feature tracking and optical flow. The main idea is to represent a video with a set of particles that move across multiple frames, and leverage long-range temporal priors while tracking the particles. 

\begin{figure*}[t]
 \includegraphics[width=\linewidth]{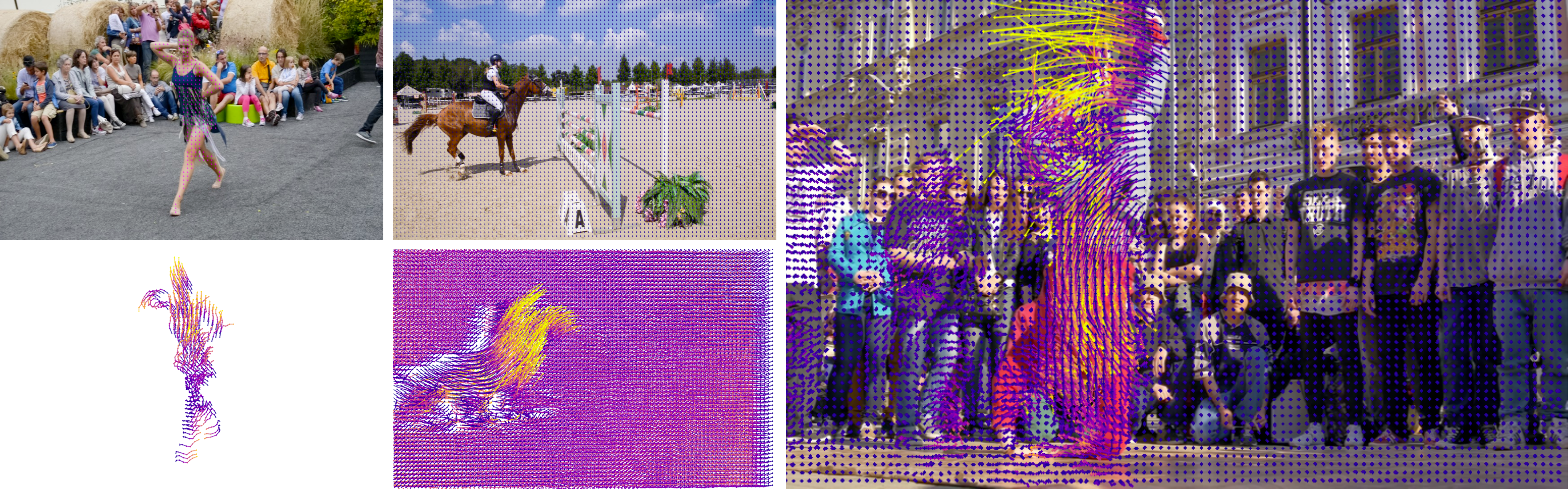}
 \caption{\textbf{Persistent Independent Particles.} Our method takes an RGB video as input, and estimates trajectories for any number of target pixels. Left: targets and their trajectories, shown separately. Right: trajectories overlaid on the pixels.} 
 \label{fig:inputsandoutputs}
\end{figure*}

Methods for feature tracking and optical flow estimation have greatly advanced since that time, but there has been relatively little work on estimating long-range trajectories at the pixel level. 
Feature correspondence methods~\cite{timecycle,dino} currently work by matching the features of each new frame to the features of one or more source frames~\cite{mast}, without taking into account temporal context. 
Optical flow methods today produce such exceedingly-accurate estimates within pairs of frames~\cite{raft} that the motion vectors can often be chained across time without much accumulation of error, but as soon as the target is occluded, it is no longer represented in the flow field, and tracking fails. 

Particle videos have the potential to capture two key elements missing from feature-matching and optical flow: (1) persistence through occlusions, and (2) multi-frame temporal context. 
If we attend to a pixel that corresponds to a point on the world surface, we should expect that point to exist across time, even if appearance and position and visibility all vary somewhat unpredictably. 
Temporal context is of course widely known to be relevant for flow-based methods, but prior efforts to take multi-frame context into account have yielded only small gains. 
Flow-based methods mainly use consecutive pairs of frames, and occasionally leverage time with a simple constant-velocity prior, which weakly conditions the current flow estimate on previous frames' flow \cite{raft,ren2018fusion}.  

We propose Persistent Independent Particles (PIPs), a new particle video method, which takes a $T$-frame RGB video as input, along with the $(x,y)$ coordinate of a target to track, and produces a $T \times 2$ matrix as output, representing the positions of the target across the given frames. The model can be queried for any number of particles, at any positions within the first frame's pixel grid. A defining feature of our approach, which differentiates it from both the original particle video algorithm and modern flow algorithms, is that it makes an extreme trade-off  between spatial awareness and temporal awareness. \textit{Our model estimates the trajectory of every target independently.} Computation is shared between particles within a video, which makes inference fast, but each particle produces its own trajectory, without inspecting the trajectories of its neighbors. This extreme choice allows us to devote the majority of parameters into a module that simultaneously learns (1) temporal priors, and (2) an iterative inference mechanism that searches for the target pixel's location in all input frames. The value of the temporal prior is that it allows the model to \textit{fail} its correspondence task at multiple intermediate frames. As long as the pixel is ``found'' at some sparse timesteps within the considered temporal span, the model can use its prior to estimate plausible positions for the remaining timesteps. 
This is helpful because appearance-based correspondence is impossible in some frames, due to occlusions, moving out-of-bounds, or difficult lighting.

We train our model entirely in synthetic data, which we call FlyingThings++, based on the FlyingThings~\cite{flyingthings16} optical flow dataset. Our dataset includes multi-frame amodal trajectories, with challenging synthetic occlusions caused by moving objects. 
In our experiments on both synthetic and real video data, we demonstrate that our particle trajectories are more robust to occlusions than flow trajectories---they can pick up an entity upon re-appearance---and also provide smoother and finer-grained correspondences than current feature-matching methods, thanks to its temporal prior. 
We also propose a method to link the model's moderate-length trajectories into arbitrarily-long trajectories, relying on a simultaneously-estimated visibility cue. Figure~\ref{fig:inputsandoutputs} displays sample outputs of our model on RGB videos from the DAVIS benchmark~\cite{davis2017}. Our code and data are publicly available at \url{https://particle-video-revisited.github.io}. 

\section{Related  Work} \label{sec:related}

\subsection{Optical Flow}

Many classic methods track points independently \cite{lucas1981iterative,tomasi1991detection}, and such point tracks see wide use in structure-from-motion \cite{bregler_recover,kong_deepnrsfm,novotny2019c3dpo} and simultaneous localization and mapping systems \cite{taketomi2017visual}. While earlier optical flow methods use optimization techniques to estimate motion fields between two consecutive frames \cite{Brox2011LargeDO,brox_densepoint}, recent methods learn such displacement fields supervised from synthetic datasets \cite{flownet,flownet2}. 
Many recent works use iterative refinements for flow estimation by leveraging coarse-to-fine pyramids~\cite{sun2018pwc}. Instead of coarse-to-fine refinements, RAFT~\cite{raft} mimics an iterative optimization algorithm, and estimates flow through iterative updates of a high resolution flow field based on 4D correlation volumes constructed for all pairs of pixels from per-pixel features. Inspired by RAFT, we also perform iterative updates of the position estimations using correlations as an input, but unlike RAFT we additionally update features.

Ren~\textit{et al.}~\cite{ren2018fusion} propose a fusion approach for multi-frame optical flow estimation. The optical flow estimates of previous frames are used to obtain multiple candidate flow estimations for the current timestep, which are then fused into a final prediction by a learnable module. 
In contrast, our method explicitly reasons about multiframe context, and iteratively updates its estimates across all frames considered. Note that without using multiple frames, it is impossible to recover an entity after occlusion. 
Janai et al.~\cite{Janai2018ECCV} is closer to our method, since it uses 3 frames as multiframe context, and explicitly reasons about occlusions. 
That work uses a constant velocity prior \cite{salgado2007temporal} to estimate motion during occlusion. In contrast, we devote a large part of the model capacity to learning an accurate temporal prior, and iteratively updates its estimates across all frames considered, in search of the object's re-emergence from occlusion. Note that without using multiple frames, it is impossible to recover an entity after occlusion. Additionally, our model is the only work that aims to recover \textit{amodal} trajectories that do not terminate at occlusions but rather can recover and re-connect with a visual entity upon its re-appearance.

\subsection{Feature Matching}

Wang and Jabri et al.~\cite{timecycle,jabri2020walk} leverage cycle consistency of time for feature matching. This allows unsupervised learning of features by optimizing a cycle consistency loss on the feature space across multiple time steps in unlabelled videos. 
Lai et al.~\cite{Lai19,mast} and Yang et al.~\cite{yang2021selfsupervised} learn feature correspondence through optimizing a proxy reconstruction objective, where the goal is to reconstruct a target frame (color or flow) by linearly combining pixels from one or more reference frames. 
Instead of using proxy tasks, supervised approaches~\cite{jiang2021cotr,germain2021NeurHal,Wiles21,wang2020learning} directly train models using ground truth correspondences across images. Features are usually extracted per-image and a transformer-based processor locates correspondences between images. In our work, we reason about point correspondences over a long temporal horizon, incorporating motion context, instead of using pairs of frames like these works. 

\subsection{Tracking with Temporal Priors}

Our work argues for using wide temporal context to track points, but this is not new for visual tracking in general. For decades, research on object-centric trackers has dealt with occlusions~\cite{zhao2004tracking} and appearance changes~\cite{matthews2004template}, and non-linear pixel-space temporal priors \cite{sidenbladh2000stochastic}, similar to our work. 
%
Here, we merely aim to bring the power of these object-tracking techniques down to the point level.

\section{Persistent Independent Particles (PIPs)}\label{sec:method}

\subsection{Setup and Overview}

Our work revisits the classic Particle Video~\cite{particlevideo} problem with a new algorithm, which we call Persistent Independent Particles (PIPs).\footnote{The countable dots on playing cards, dice, or dominoes are also called ``pips''.}  
We take as input an RGB video with $T$ frames, and the $(x_1,y_1)$ coordinate of a pixel on the first frame, indicating the target to track. As output, we produce per-timestep coordinates $(x_t,y_t)$ tracking the target across time, and per-timestep visibility/occlusion estimates $v_t \in [0,1]$. The model can be queried for $N$ target points in parallel, and some computation will be shared between them, but the model does not share information between the targets' trajectories. 



At training time, we query the model with points for which we have ground-truth trajectories and visibility labels. We supervise the model's $(x_t,y_t)$ outputs with a regression objective, and supervise $v_t$ with a classification objective. 
At test time, the model can be queried for the trajectories of any number of points. 

We use the words ``point'' and ``particle'' interchangeably to mean the things we are tracking, and use the word ``pixel'' more broadly to indicate any discrete cell on the image grid. Note that although the tracking targets are specified with single pixel coordinates, tracking successfully requires (at least) taking into account the local spatial context around the specified pixel. 

\begin{figure*}[t]
 \includegraphics[width=\linewidth]{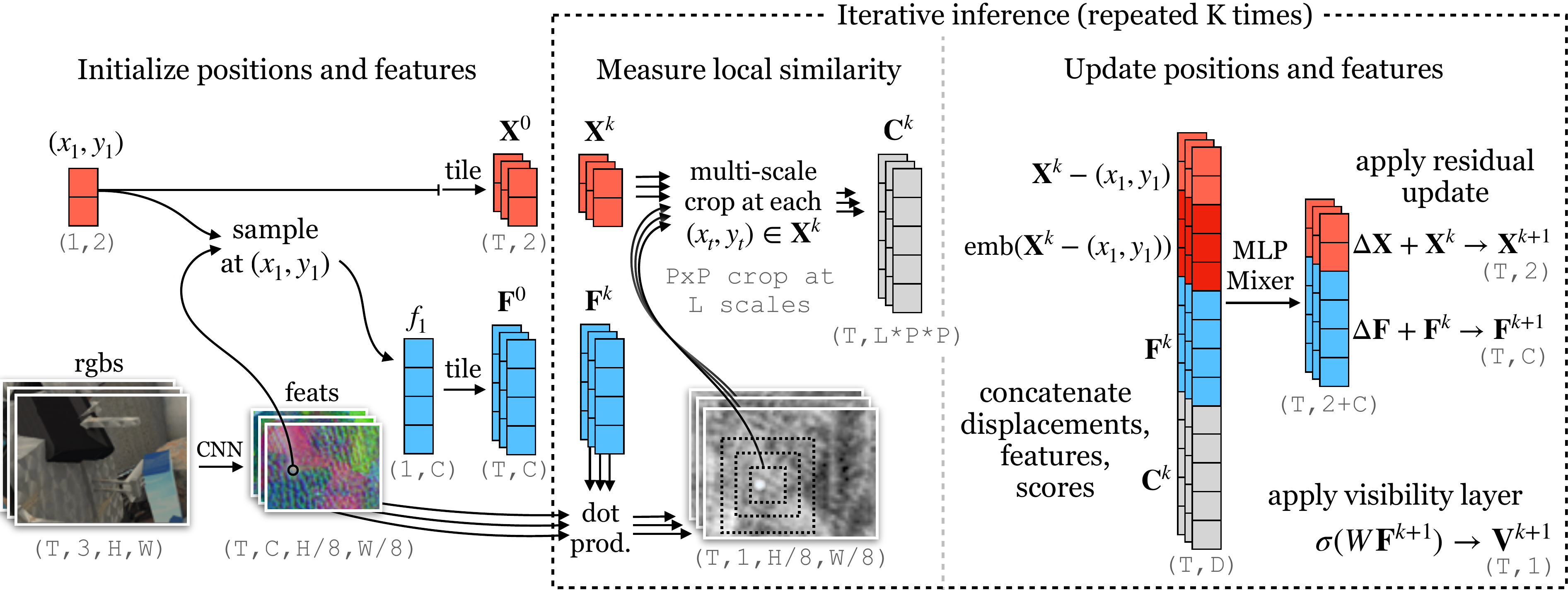}
 \caption{\textbf{Persistent Independent Particles (PIPs) architecture.} Given an RGB video as input, along with a location in the first frame indicating what to track, our model initializes a multi-frame trajectory, then computes features and correlation maps, and iteratively updates the trajectory and its corresponding sequence of features, with a deep MLP-Mixer model. From the computed features, the model also estimates a visibility score for each timestep of the trajectory.} 
 \label{fig:overview}
\end{figure*}

Our overall approach has four stages, somewhat similar to the RAFT optical flow method \cite{raft}: extracting visual features (Section \ref{sec:st1}),  initializing a list of positions and features for each target (Section \ref{sec:st2}),  locally measuring appearance similarity (Section \ref{sec:st3}), and  repeatedly updating the positions and features for each target (Section \ref{sec:st4}).  Figure~\ref{fig:overview} shows an overview of the method. 

\subsection{Extracting Features} \label{sec:st1}

We begin by extracting features from every frame of the input video. In this step, each frame is processed independently with a 2D convolutional network (i.e., no temporal convolutions). The network produces features at 1/8 resolution.




\subsection{Initializing Each Target}\label{sec:st2}

After computing feature maps for the video frames, we compute a feature vector for the target, by bilinearly sampling inside the first feature map at the first (given) coordinate, obtaining a feature vector $f_1$, with $C$ channels. We use this sampled feature to initialize a trajectory of features, by simply tiling the feature across time, yielding a matrix $\mathbf{F}^{0}$ sized $T \times C$. This initialization implies an appearance constancy prior. 

We initialize the target's trajectory of positions in a similar way. We simply copy the initial position across time, yielding a matrix $\mathbf{X}^0$, shaped $T \times 2$. 
This initialization implies a zero-velocity prior, which essentially assumes nothing about the target's motion.

During inference, we will update the trajectory of features, tracking appearance changes, and update the trajectory of positions, tracking motion. We use the superscript $k$ to indicate the update iteration count, as in $\mathbf{X}^k, \mathbf{F}^k$. 

\subsection{Measuring Local Appearance Similarity} \label{sec:st3}

\looseness=-1 We would like to measure how well our trajectory of positions, and associated trajectory of features, matches with the pre-computed feature maps.
We compute visual similarity maps by correlating each feature $f_t$ in $\mathbf{F}^k$ with the feature map of the corresponding timestep, and then obtain ``local'' scores by bilinearly sampling a crop centered at the corresponding position $(x_t,y_t)$. 
This step returns patches of un-normalized similarity scores, where large positive values indicate high similarity between the target's feature and the convolutional features at this location. 
The sequence of score patches is shaped $T \times P \cdot P$, where $P$ is the size of the patch extracted from each correlation map. 
Similar to RAFT~\cite{raft}, we find it is beneficial to create a spatial pyramid of these score patches, to obtain similarity measurements at multiple scales. We denote our set of multi-scale score crops $\mathbf{C}^k$, shaped $T \times P \cdot P \cdot L$, where $L$ is the number of levels in the pyramid.  


\subsection{Iterative Updates} \label{sec:st4}


The main inference step for our model involves updating the sequence of positions, and updating the sequence of features. 
To perform this update, we take into account all of the information we have computed thus far: the feature matrix $\mathbf{F}^k$, the correlation matrix $\mathbf{C}^k$, and displacements computed from the position matrix $\mathbf{X}^k$. 
To compute displacements from $\mathbf{X}^k$, we subtract the given position $(x_1,y_1)$ from each element of the matrix. 
Using displacements instead of absolute positions makes all input trajectories appear to start at $(0, 0)$, which makes our model translation-invariant. To make the displacements easier to process by the model, we employ sinusoidal position encodings \cite{vaswani2017attention}, motivated by the success of these encodings in vision transformers \cite{dosovitskiy2020image}.


We concatenate this broad set of inputs on the channel dimension, yielding a new matrix shaped $T \times D$, and process them with a 12-block MLP-Mixer \cite{mlpmixer}, which is a parameter-efficient all-MLP architecture with design similarities to a transformer. As output, this module produces updates for the sequence of positions and sequence of features, $\Delta\mathbf{X}$ and $\Delta\mathbf{F}$, which we apply with addition: $\mathbf{F}^{k+1} = \mathbf{F}^k + \Delta\mathbf{F}$, and  $\mathbf{X}^{k+1} = \mathbf{X}^k + \Delta\mathbf{X}.$
After each update, we compute new correlation pyramids at the updated coordinates, using the updated features. 

The update module is iterated $K$ times. After the last update, the positions $\mathbf{X}^K$ are treated as the final trajectory, and the features $\mathbf{F}^K$ are sent to a linear layer and sigmoid, to estimate per-timestep visibility scores $\mathbf{V}^K$. 


\subsection{Supervision} 

We supervise the model using the $L_1$ distance between the ground-truth trajectory and the estimated trajectory (across iterative updates), with exponentially increasing weights, similar to RAFT \cite{raft}: 
\begin{equation}
    \mathcal{L}_{\textrm{main}} = \sum_k^K \gamma^{K-k} || \mathbf{X}^k - \mathbf{X}^* ||_1,
\end{equation}
where $K$ is the number of iterative updates, and we set $\gamma=0.8$. Note that this loss is applied even when the target is occluded, or out-of-bounds, which is possible since we are using synthetically-generated ground truth. This is the main loss of the model, and the model can technically train using only this, although it will not learn visibility estimation and convergence will be slow. 

On the model's visibility estimates, we apply a cross entropy loss: 
\begin{equation}
\mathcal{L}_{\textrm{ce}} = \mathbf{V}^* \log \mathbf{V} + (1 -  \mathbf{V}^*)\log (1 - \mathbf{V}).
\end{equation}

We find it accelerates convergence to directly supervise the score maps to peak in the correct location (i.e., the location of the true correspondence):
\begin{equation}
\mathcal{L}_{\textrm{score}} = -\log ( \exp(c_i)/\sum_j{\exp(c_j})) 1 \{ \mathbf{V}^* \neq 0 \},
\end{equation}
where $c_j$ represents the match score at pixel $j$, and $i$ is pixel index with the true correspondence, and $1\{\mathrm{V}^*\neq 0\}$ selects indices where the target is visible. 



\subsection{Test-Time Trajectory Linking}


At test time, it is often desirable to generate correspondences over longer timespans than the training sequence length $T$. To generate these longer trajectories, we may repeat inference starting from any timestep along the estimated trajectory, treating $(x_t,y_t)$ as the new $(x_1,y_1)$, and thereby ``continuing'' the trajectory up to $(x_{t+T}, y_{t+T})$. However, doing this naively (e.g., always continuing from the last timestep), can quickly cause tracking to drift. In particular, it is crucial to avoid continuing the trajectory from a timestep where the target is occluded. Otherwise, the model will switch to tracking the occluder. To avoid these identity switches, we make use of our visibility estimates, and seek a late timestep whose visibility score is high. This allows the model to skip past frames where the target was momentarily occluded, as long as the temporal span of the occlusion is less than the temporal span of the model ($T$). 
We initialize a visibility threshold conservatively at 0.99, and decrease it in increments of 0.01 until a valid selection is found.  To lock the model into tracking the ``original'' target, we simply re-use the original $\mathbf{F}^0$ across all re-initializations. 

\section{Implementation Details}

\hspace{\parindent}\textbf{CNN.} Our CNN uses the ``BasicEncoder'' architecture from the official RAFT codebase \cite{raftcode}. This architecture has a $7\times7$ convolution with stride 2, then 6 residual blocks with kernel size $3 \times 3$, then a final convolution with kernel size $1 \times 1$. The CNN has an output dimension of $C=256$. 

\textbf{Local correlation pyramids.} We use four levels in our correlation pyramids, with radius 3, yielding four $7 \times 7$ correlation patches per timestep.

\textbf{MLP-Mixer.} The input to the MLP-Mixer is a sequence of displacements, features, and correlation pyramids. 
The per-timestep inputs are flattened, then treated as a sequence of vectors (i.e., ``tokens'') for the MLP-Mixer. We use the MLP-Mixer architecture exactly as described in the original paper; at the end of the model there is a mean over the sequence dimension, followed by a linear layer that maps to a channel size of $T \cdot (C + 2)$. 

\textbf{Updates.} We reshape the MLP-Mixer's outputs into a sequence of feature updates and a sequence of coordinate updates, and apply them with addition. 
We train and test with 6 updates. 

\textbf{Visibility.} We use a linear layer to map the last update iteration's pixel-level feature sequence into visibility logits. 

\textbf{Training.} We train with a batch size of 4, distributed across four GPUs. At training time, we use a resolution of $368 \times 512$. For each element of the batch, (after applying data augmentation,) we randomly sample 128 trajectories which begin in-bounds and un-occluded. We train for 100,000 steps, with a learning rate of 3e-4 with a 1-cycle schedule \cite{smith2019super}, using the AdamW optimizer. Training takes approximately 2 days on four GeForce RTX 2080s.

\textbf{Hyperparameters.} We use $T=8$ (timesteps considered simultaneously by the model), and $K=6$ (update iterations). The model can in general be trained for any $T$, but we found that the model was more difficult to train at $T=32$, likely because the complexity of trajectories grows rapidly with their length under our model, as there is no weight sharing across time. 
On the other hand, the temporal sensitivity allows our model to learn more complex temporal priors. 
We found that $K>6$ performs similar to $K=6$. 
Although we train the model at a spatial stride of 8, this may be changed at test time; we find that a stride of 4 works best on our (high-resolution) test datasets. 

\looseness=-1 \textbf{Complexity.} \textit{Speed:} 
When the number of targets is small enough to fit on a GPU (e.g., 256 targets for a 12G GPU), our model is faster than RAFT (200ms vs. 2000ms at $480\times1024$). RAFT is comparatively slow because (1) it is too memory-heavy to compute all frames' flows in parallel, so we must run it $T-1$ times, and (2) it attempts to track all pixels instead of a given set of targets. 
When the number of targets is too large to fit on a GPU (in parallel), our model processes them in batches, and in this case PIPs may be slower than RAFT.
\textit{Memory:} 
Our model's memory scales primarily with $T\cdot N$, where $N$ is the number of particles being tracked, due to the iterated MLP-Mixer which consumes a $T$-length sequence of features per particle. 



\section{Experiments}\label{sec:exp}

We train our model in a modified version of FlyingThings~\cite{flyingthings16}, which we name FlyingThings++ (discussed more below).
We evaluate our model on tracking objects in FlyingThings++, tracking vehicles and pedestrians in KITTI~\cite{kitti}, tracking heads in a crowd in CroHD~\cite{sundararaman2021tracking}, and finally, propagating keypoints in animal videos in BADJA~\cite{badja}. 
We visualize trajectory estimates in DAVIS videos in Figure~\ref{fig:inputsandoutputs}, to illustrate the method's generality, and visualize the estimates against ground truth in Figures~\ref{fig:alldat_qual} and~\ref{fig:badja_qual}. 
In the supplementary we include video visualizations of our results. All of our experiments evaluate the same PIP model---we do not customize any parameters for the individual test domains. 

\subsection{Training Data: FlyingThings++}

To train our model, we created a synthetic dataset called FlyingThings++, based on FlyingThings~\cite{flyingthings16}. The original FlyingThings is typically used in combination with other flow datasets to train optical flow models. We chose FlyingThings because (1) its visuals and motions are extremely complex, which gives hope of generalizing to other data, and (2) it provides 10-frame videos with ground-truth forward and backward optical flow, and instance masks, from which we can mine accurate multi-frame trajectories. 

To create multi-frame trajectories, we chain the flows forward, and then discard chains which (i) fail a forward-backward consistency check, (ii) leave the image bounds, or (iii) shift from one instance ID to another. This leaves a sparse set of 8-frame trajectories, which cover approximately 30\% of the pixels of the first frame. These checks ensure that the trajectories are accurate, but leave us with a library of trajectories where the target is visible on every timestep. Therefore, it is necessary to add \textit{new occlusions} on top of the video. We do this on-the-fly during batching: for each FlyingThings video in the batch, we randomly sample an object from an alternate FlyingThings video, paste it directly on top of the current video, overwriting the pixels within its mask on each frame. We then update the ground-truth to reflect the occluded area on each frame, as well as update the trajectory list to include the trajectories of the added object. 

Combining all videos with at least 256 valid 8-frame trajectories, we obtain a total of 13085 training videos, and 2542 test videos. To expand the breadth of the training set, we augment the data on-the-fly with random color and brightness changes, random scale changes, crops which randomly shift across time, random Gaussian blur, and random horizontal and vertical flips. 

\subsection{Baselines}
In our experiments we consider the following baselines.

\textbf{Recurrent All-Pairs Field Transforms (RAFT)} \cite{raft} represents the state-of-the-art in optical flow estimation, where a high resolution flow field is refined through iterative updates, based on lookups from a 4D cost volume constructed between all pairs of pixels. Similar to our method, RAFT has been trained on FlyingThings (including occlusions and out-of-bounds motions), but only has a 2-frame temporal span. To generate multi-frame trajectories with RAFT at test time, we compute flow with all consecutive pairs of frames, and then compute flow chains at the pixels queried on the first frame. To continue chains that travel out of bounds, we clamp the coordinates to the image bounds and sample at the edge of the flow map.

\textbf{DINO} \cite{dino} is a vision transformer (ViT-S \cite{dosovitskiy2020image} with patch size 8) trained on ImageNet with a self-supervision objective based on a knowledge distillation setup that builds invariance to image augmentations. 
To use this model for multi-frame correspondence, we use the original work's code for instance tracking, which uses nearest neighbor between the initial frame and the current frame, as well as nearest-neighbor between consecutive frames, and a strategy to restrict matches to a local neighborhood around previous matches. We report results with and without this ``windowing'' strategy. 


\textbf{Contrastive Random Walk (CRW)} \cite{jabri2020walk} treats the video as a space-time graph, with edges containing transition probabilities of a random walk, and computes long-range correspondences by walking across the graph.
The model learns correspondences between pixels from different frames by optimizing an objective that encourages correspondences to be cycle-consistent across time (i.e., forward-backward consistency), including across frame skips. This method tracks in a similar way to DINO. 

\textbf{Memory-Augmented Self-supervised Tracker (MAST)} \cite{mast} learns correspondences between features by reconstructing the target frame with linear combinations of reference frames. At test time the correspondences are predicted autoregressively. The model is trained on OxUvA~\cite{valmadre2018long} and YouTube-VOS~\cite{xu2018youtube}

\textbf{Video Frame-level Similarity (VFS)} \cite{xu2021vfs} learns an encoder that produces frame-level embeddings which are similar within a video, and dissimilar across videos. This model is trained on Kinetics-400~\cite{kay2017kinetics}. 
    
\textbf{ImageNet ResNet} \cite{he2016deep} is a ResNet50 supervised for classification with ImageNet labels, and evaluated the same way as DINO. 


\subsection{Trajectory Estimation in FlyingThings++}


Using 8-frame videos from the FlyingThings++ test set as input, we estimate trajectories for all pixels for which we have ground-truth, and evaluate the average distance between each estimated trajectory and its corresponding ground truth, averaging over all 8 timesteps. We are especially interested in measuring how our model's performance compares with the baselines when the target gets occluded or flies out of frame. We use a crop size of $384 \times 512$, which puts many trajectories flying out of image bounds. For this evaluation, since occlusions are extremely common, we count a trajectory as ``visible'' if at least half of its timesteps are visible, and count it as ``occluded'' otherwise. 

We compare our model against DINO \cite{dino}, representing the state-of-the-art for feature matching, and RAFT \cite{raft}, representing the state-of-the-art for flow. Table~\ref{tab:flt_test} shows the results across the different evaluations on the test set. DINO struggles overall, likely because of the rapid motions, and performs worse on occluded pixels than on visible ones. This makes sense because during occlusions DINO cannot make any matches. RAFT obtains reasonable accuracy for visible pixels (considering the speed of the motion in this data), but its errors drastically increase for the heavily-occluded trajectories. We have also re-trained RAFT in this data and its performance did not improve, likely because these are \textit{multi-timestep} occlusions, which chaining-based methods cannot accommodate. 
Inspecting the results manually, we see that RAFT's trajectories often drift off the targets and follow the occluders, which makes sense because during occluded timesteps the flow field does not contain the targets. 
Our model, in contrast, is able to locate the targets after they re-emerge from occlusions, and inpaint the missing portions of the trajectories, leading to better performance overall. 


\subsection{Trajectory Estimation in KITTI}

We evaluate on an 8-frame point trajectory dataset that we created from the ``tracking'' subset of the KITTI~\cite{kitti} urban scenes benchmark. 
The data is at 10 FPS, and we use this framerate as-is. We use videos from sequences 0000-0009, which include mostly vehicles, as well as 0017 and 0019, which include mostly pedestrians. To create 8-frame trajectories, we sample a 3D box annotation that has at least 8 valid timesteps, compute the mean LiDAR point within the box on the first timestep, transform it in 3D to its corresponding location on every other step, and project this location into pixel coordinates. 
We approximate visibility/occlusion labels in this data by checking if another object crosses in front of the target. We resize the frames to $320 \times 512$. 


Table~\ref{tab:kitti_test} shows the results. RAFT performs slightly better than PIPs for targets that stay visible, but PIPs performs slightly better for targets that undergo occlusions. DINO's error is much higher. In this data, the motions are relatively slow, and we find that DINO's trajectories are only coarsely tracking the targets, likely because of the low-resolution features in that model. 
Qualitative results for our model are shown in Figure~\ref{fig:alldat_qual}-middle. 

\begin{table}[t]
\parbox{.3\linewidth}{
\centering
\caption{{Trajectory error in FlyingThings++.}
PIP trajectories are more robust to occlusions.}\label{tab:flt_test}
\begin{tabular}{lcc}
\toprule
{Method} & {Vis.} & {Occ.} \\
\midrule
DINO~\cite{dino} & 40.68 & 77.76 \\
RAFT~\cite{raft} & 24.32 & 46.73\\
PIPs (ours) & \textbf{15.54} & \textbf{36.67}\\
\bottomrule
\end{tabular}
}
\hfill
\parbox{.3\linewidth}{
\centering
\caption{{Trajectory error in KITTI.} PIP and RAFT trajectories are similar; DINO lags behind both.}\label{tab:kitti_test}
\begin{tabular}{lcc}
\toprule
{Method} & {Vis.} & {Occ.}\\
\midrule
DINO~\cite{dino} & 13.33 & 13.45 \\
RAFT~\cite{raft} & \textbf{4.03} & 6.79 \\
PIPs (ours) & 4.40 & \textbf{5.56}\\
\bottomrule
\end{tabular}
}
\hfill
\parbox{.3\linewidth}{
\centering
\caption{{{Trajectory error in CroHD.} PIP trajectories achieve better accuracy overall.}\label{tab:heads}}
\begin{tabular}{lcc}
\toprule
{Method} & {Vis.} & {Occ.}\\
\midrule
DINO~\cite{dino} &  22.50 & 26.06\\
RAFT~\cite{raft} & 7.91 & 13.04 \\
PIPs (ours) & \textbf{5.16} & \textbf{7.56}\\
\bottomrule
\end{tabular}
}
\end{table}

\subsection{Trajectory estimation in CroHD}

\begin{figure*}[t]
\begin{center}
 \includegraphics[width=1.0\linewidth]{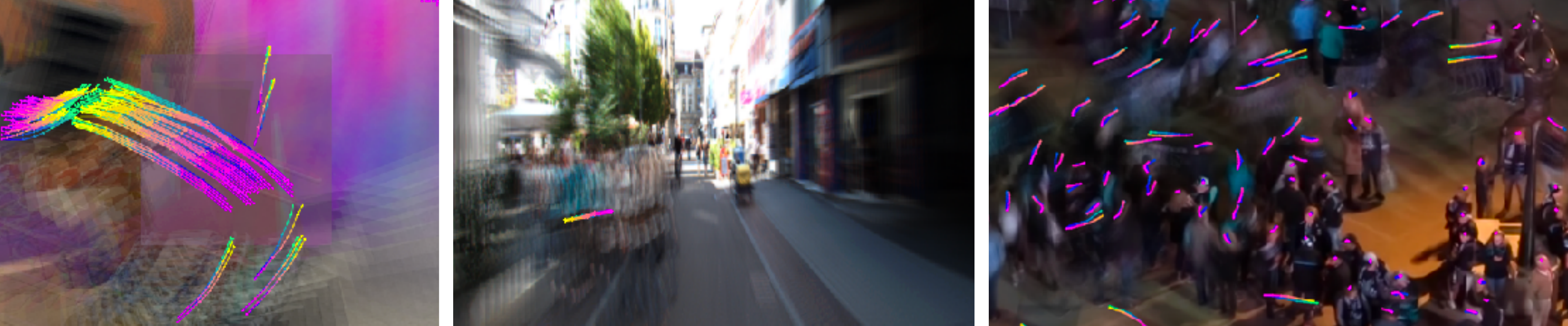}
 \end{center}
 \caption{\textbf{Qualitative results in FlyingThings++ (left), KITTI (middle), and CroHD (right).} We visualize a video with the mean of its RGB. We trace the PIP trajectories in pink-to-yellow, and show ground truth in blue-to-green. FlyingThings++ is chaotic, but training on this data allows our model to generalize.}\label{fig:alldat_qual}
\end{figure*}

We evaluate on the Crowd of Heads Dataset (CroHD)~\cite{sundararaman2021tracking}, which consists of high-resolution ($1080 \times 1920$) videos of crowds, with annotations tracking the heads of people in the crowd. We evaluate on 8-frame sequences extracted from the dataset.
We subsample the frames in the original dataset so that the FPS is reduced by a factor of 3.
We use a resolution of $768 \times 1280$ for PIPs and RAFT, and $512 \times 768$ for DINO (since otherwise it exceeds our 24G GPU memory).  We filter out targets whose motion is below a threshold distance, and split the evaluation between targets that stay visible and those that undergo occlusions (according to the visibility annotations in the data). The results are shown in Table~\ref{tab:heads}, and Figure~\ref{fig:alldat_qual}-right. In this data, PIP trajectories outperform RAFT and DINO by a wide margin, in both visibility conditions. 
In this data, DINO likely struggles because the targets are all so similar to one another. 

\begin{table*}[t]
\begin{center}
\caption{\textbf{PCK-T in BADJA.} In this evaluation, keypoints are initialized in the first frame of the video, and are propagated to the end of the video; PCK-T measures the accuracy of this propagation. In each column, we bold the best result, and underline the second-best. Above the middle bar, we give methods a spatial window (marked ``Win.'') to constrain how they propagate labels,  which is a common strategy in existing work. 
Our method wins in most videos, but DINO performs well also.}\label{tab:badja_test}
\begin{tabular}{lccccccc|c}
\toprule
{Method} & bear & camel & cows & dog-a & dog & horse-h & horse-l & Avg. \\
\midrule
Win. DINO \cite{dino} & \textbf{77.9} & 69.8 & \textbf{83.7} & \underline{17.2} & \underline{46.0} & 29.1 & 50.8 & \underline{53.5}\\
Win. ImageNet ResNet \cite{he2016deep} & 70.7 & 65.3 & 71.7 & 6.9 & 27.6 & 20.5 & 49.7 & 44.6 \\
Win. CRW \cite{jabri2020walk} & 63.2 & \underline{75.9} & 77.0 & 6.9 & 32.8 & 20.5 & 22.0 & 42.6 \\
Win. VFS \cite{xu2021vfs} & 63.9 & 74.6 & 76.2 & 6.9 & 35.1 & 27.2 & 40.3 & 46.3 \\
Win. MAST \cite{mast} & 35.7 & 39.5 & 42.0 & 10.3 & 8.6 & 12.6 & 14.7 & 23.3\\
Win. RAFT \cite{raft} & 64.6 & 65.6 & 69.5 & 3.4 & 38.5 &
33.8 & 28.8 & 43.5\\
\midrule
DINO \cite{dino} & 75.0 & 59.2 & 70.6 & 10.3 & \textbf{47.1} & 35.1 & \underline{56.0} & 50.5 \\
ImageNet ResNet \cite{he2016deep} & 65.4 & 53.4 & 52.4 & 0.0 & 23.0 & 19.2 & 27.2 & 34.4 \\
CRW \cite{jabri2020walk} & 66.1 & 67.2 & 64.7 & 6.9 & 33.9 & 25.8 & 27.2 & 41.7 \\
VFS \cite{xu2021vfs} & 64.3 & 62.7 & 71.9 & 10.3 & 35.6 & 33.8 & 33.5 & 44.6 \\
MAST \cite{mast} & 51.8 & 52.0 & 57.5 & 3.4 & 5.7 & 7.3 & 34.0 & 30.2 \\
RAFT \cite{raft} & 64.6 & 65.6 & 69.5 & 13.8 & 39.1 & \underline{37.1} & 29.3 & 45.6 \\
PIPs (ours) & \underline{76.3} & \textbf{81.6} & \underline{83.2} & \textbf{34.2} & 44.0 & \textbf{57.4} & \textbf{59.5} & \textbf{62.3} \\
\bottomrule
\end{tabular}
\end{center}
\end{table*}

\begin{figure*}[t]
\begin{center}
 \includegraphics[width=1.0\linewidth]{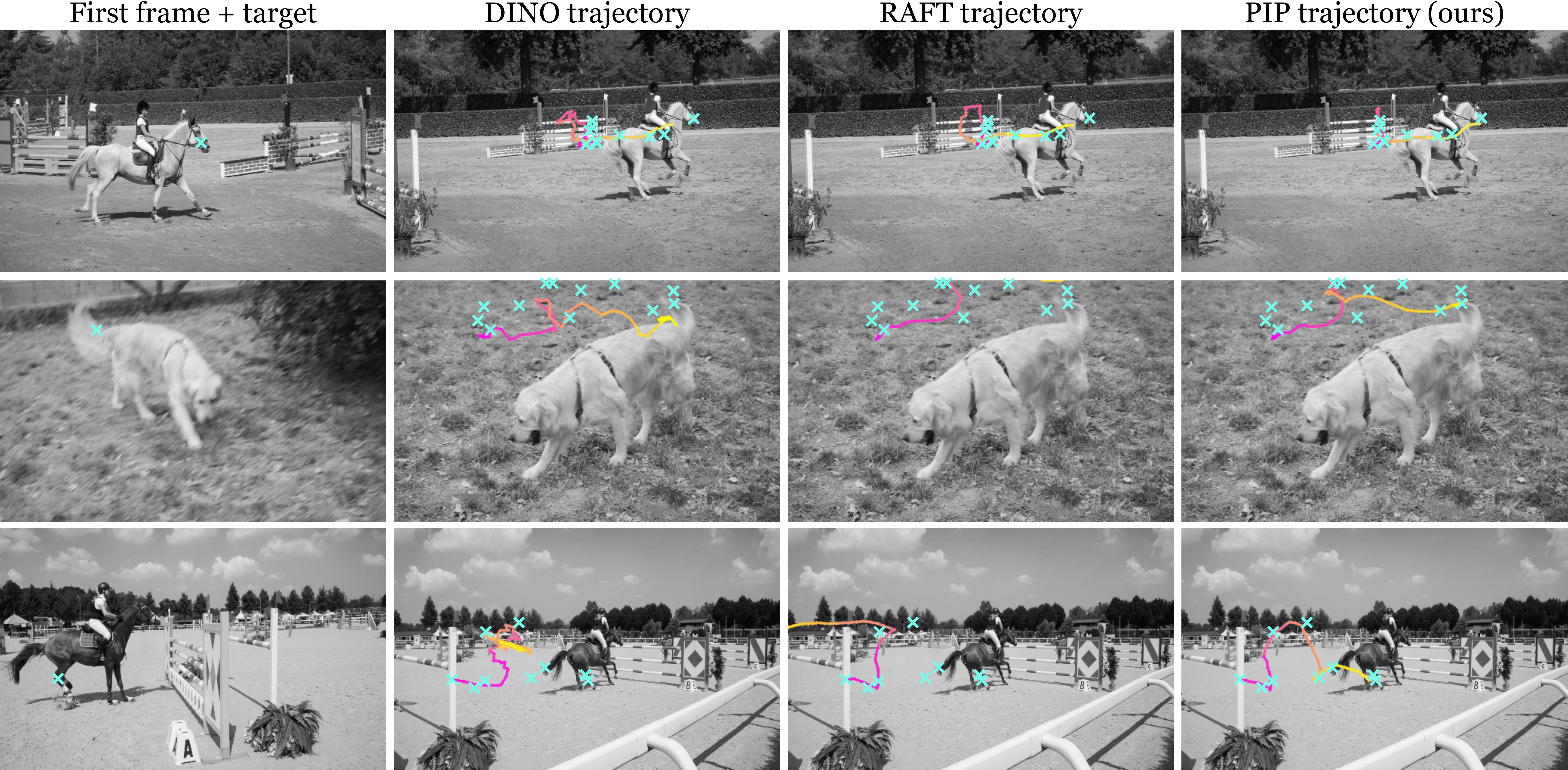}
 \end{center}
 \caption{\textbf{Comparison with baselines in BADJA, on videos with occlusions.} For each method, we trace the estimated trajectory with a pink-to-yellow colormap. The sparse ground truth is visualized with cyan~$\times$~marks. 
 In the first video, the target (on the dog's tail) leaves the image bounds then returns into view. In the second video, the target (on the horse's leg) is momentarily occluded, causing RAFT to lose track entirely. 
For a more detailed view, please watch the supplementary video. 
} \label{fig:badja_qual}
\end{figure*}

\subsection{Keypoint Propagation in BADJA}

\looseness=-1 BADJA~\cite{badja} is a dataset of animal videos with keypoint annotations. These videos overlap with the DAVIS dataset~\cite{davis2017}, but include keypoint annotations. 
Keypoint annotations exist on approximately 1/5 frames, and the standard evaluation is Percentage of Correct Keypoint-Transfer (PCK-T), where keypoints are provided on a reference image, and the goal is to propagate these annotations to other frames. A keypoint transfer is considered correct if it is within a distance of $0.2 \sqrt{A}$ from the true pixel coordinate, where $A$ is the area of the ground-truth segmentation mask on the frame. We note that some existing methods test on a simplified version of this keypoint propagation task, where the ground-truth segmentation is available on every frame of the video \cite{yang2021lasr,yang2021viser}. Here, we focus on the harder setting, where the ground-truth mask is unknown. We also note that feature-matching methods often constrain their correspondences to a local spatial window around the previous frame's match \cite{dino}. We therefore report additional baseline results using the qualifier ``Windowed'', but we focus PIPs on the un-constrained version of the problem, where keypoints need to be propagated from frame 1 to every other frame, with no other knowledge about motion or position. We test at a resolution of $320 \times 512$. We have observed that higher accuracy is possible (for many methods) at higher resolution, but this puts some models beyond the memory capacity of our GPUs.

Table~\ref{tab:badja_test} shows the results of the BADJA evaluation. On four of the seven videos, our model produces the best keypoint tracking accuracy, as well as the best on average by a margin of 9 points. DINO \cite{dino} obtains the best accuracy in the remaining videos, though its widest margin over our model is just 3 points. Interestingly, windowing helps DINO (and other baselines) in some videos but not in others, perhaps because of the types of motions in DAVIS.  
We note that DAVIS has an object-centric bias (i.e., the target usually stays near the center of the frame), which translation-sensitive methods like DINO can exploit, since their features encode image position embeddings; RAFT and PIPs track more generally. In Figure~\ref{fig:badja_qual} we visualize trajectories on targets that undergo momentary occlusions, illustrating how DINO tracks only coarsely, and how RAFT loses track after the occlusion, while PIPs stay on target. 

\subsection{Limitations}\label{sec:limitations}
Our model has two main limitations. First is our unique extreme tradeoff, of spatial awareness for temporal awareness. Although this maximizes the power of the temporal prior in the model, it discards information that could be shared between trajectories. 
We are indeed surprised that single-particle tracking performs as well as it does, considering that spatial smoothness is known to be essential for accurate optical flow estimation. Extending our architecture to concurrent estimation of multiple point trajectories is a direct avenue for future work.

Our second main limitation stems from the MLP-Mixer. Due to this architecture choice, our model is not recurrent across time. Although longer trajectories can be produced by re-initializing our inference at the tail of an initial trajectory, our model will lose the target if it stays occluded beyond the model's temporal window. We have tried models that are convolutional across time, and that use self-attention across the sequence length, but these did not not perform as well as the MLP-Mixer on our FlyingThings++ tests. Taking advantage of longer and potentially varying temporal context would help the model track through longer periods of ambiguity, and potentially leverage longer-range temporal priors.



\section{Conclusion} \label{sec:conclusion}
We propose Persistent Independent Particles (PIPs), a method for multi-frame point trajectory estimation through occlusions. Our method combines cost volumes and iterative inference with a deep temporal network, which jointly reasons about location and appearance of visual entities across multiple timesteps. We argue that optical flow, particle videos, and feature matches cover different areas in the spectrum of pixel-level correspondence tasks. 
Particle videos benefit from temporal context, which matching-based methods lack, and can also survive multi-frame occlusions, which is missing in flow-based methods. 
Given how tremendously useful optical flow and feature matching have been for driving progress in video understanding, we hope the proposed multi-frame trajectories will spark interest in architectures and datasets designed for longer-range fine-grained correspondences.
\\

\textbf{Acknowledgements.} 
This material is based upon work supported by Toyota Research Institute (TRI),  US Army contract W911NF20D0002, a DARPA Young Investigator Award, an NSF CAREER award, an AFOSR Young Investigator Award,  and DARPA Machine Common Sense. Any opinions, findings and conclusions or recommendations expressed in this material are those of the authors and do not necessarily reflect the views of the United States Army or the United States Air Force.


\bibliographystyle{splncs04}
\bibliography{99_refs}


\end{document}